\documentclass[10pt,twocolumn,letterpaper]{article}

\usepackage{ijcb}
\usepackage{times}
\usepackage{epsfig}
\usepackage{graphicx}
\usepackage{amsmath}
\usepackage{amssymb}

\usepackage{xcolor}
\usepackage{tabu}

\usepackage{times}
\usepackage{epsfig}
\usepackage{graphicx}

\usepackage{caption}
\usepackage{makecell}
\usepackage{booktabs}
\usepackage{subfig}
\usepackage{multirow}

\usepackage{multicol}
\usepackage{enumitem}

\usepackage{url}

\usepackage{breakurl}
\usepackage[breaklinks]{hyperref}

\usepackage{algorithm}
\usepackage[noend]{algpseudocode}

\usepackage{adjustbox} % align graphics

\usepackage{authblk}

\ijcbfinalcopy % *** Uncomment this line for the final submission

\ifijcbfinal\fi
\begin{document}

%%%%%%%%% TITLE
\title{Beyond Identity: What Information Is Stored in Biometric Face Templates?}

\author[1,2]{Philipp Terh\"orst}
\author[1,2]{Daniel F\"ahrmann}
\author[1]{Naser Damer}
\author[1,2]{Florian Kirchbuchner}
\author[1,2]{Arjan Kuijper}
\affil[1]{Fraunhofer Institute for Computer Graphics Research IGD, 64283 Darmstadt, Germany}
\affil[2]{Department of Computer Science, Technical University of Darmstadt, 64289 Darmstadt, Germany}

\setcounter{Maxaffil}{0}
\renewcommand\Affilfont{\itshape\small}

%% omit the following lines up until the closing ``}''.
%% Additional authors and addresses can be added with ``\and'',
%% just like the second author.
%% To save space, use either the email address or home page, not both
%\and

\maketitle
\thispagestyle{empty}

%%% add later
%fusion paper RAAGE
%Works on soft-biometric priavcy

%%%%%%%%% ABSTRACT
\begin{abstract}
%%% written for a popentially interested non-experts
\vspace{-2mm}

% 1 - provide background factual information
Deeply-learned face representations enable the success of current face recognition systems.   
% 2 - mention previous work, their challenges/research gap
Despite the ability of these representations to encode the identity of an individual, recent works have shown that more information is stored within, such as demographics, image characteristics, and social traits.
%This can cause biased decisions from face recognition systems.
This threatens the user's privacy, since for many applications these templates are expected to be solely used for recognition purposes.
% 3 - combine the method, the general aim and the specific aim of the study
Knowing the encoded information in face templates helps to develop bias-mitigating and privacy-preserving face recognition technologies.
% 4 - summarises the methodology and provide details / advantages of the approach
This work aims to support the development of these two branches by analysing face templates regarding 113 attributes.
% 5 - provide also details of the experimentat setup 
Experiments were conducted on two publicly available face embeddings.
For evaluating the predictability of the attributes, we trained a massive attribute classifier that is additionally able to accurately state its prediction confidence.
This allows us to make more sophisticated statements about the attribute predictability.
% 6 - indicate the achievement of this study
The results demonstrate that up to 74 attributes can be accurately predicted from face templates.
Especially non-permanent attributes, such as age, hairstyles, haircolors, beards, and various accessories, found to be easily-predictable.
% 7 - present the implications of the study / numerical details
Since face recognition systems aim to be robust against these variations, future research might build on this work to develop more understandable privacy preserving solutions and build robust and fair face templates.
%Future work have to address this problem.
% and non-biometric systems.
\end{abstract}

\vspace{-4mm}
%%%%%%%%% BODY TEXT
\section{Introduction}
\vspace{-1mm}
%%% introduce the reader to the topic, make him curious and show what will follow

%%% Basic components:
%%% 1 - establish the importance of you field
%       provide background information
%       define the terminology of key words
%       present the problem area / current research
%%% 2 - previous and or current research 
%%% 3 - locate a gap in the research
%       describe the problem you will address
%       present a prediction to be tested
%%% 4 - describe the present paper

% 1 - etablish the importance of the research topic
The advances of deep neural representations lead to high-performing face recognition solutions  \cite{FRVT2018}.
Due to the achieved performance, face recognition systems spread world-wide and increasingly affect our daily life \cite{DBLP:conf/btas/DamerWBBT0K18}.
% 2 - provide general background information for the reader
Despite that these face representations are trained to enable recognition of individuals, previous works showed that more information than just the identity are embedded.
% 3 - do the same as in 1 and 2 but in a more specific/detailed way using research references to support both the background facts and the claim of significance
They demonstrated face templates contain information about head pose \cite{DBLP:conf/fgr/PardeCHCSCO17}, image characteristics (such as quality \cite{DBLP:journals/corr/Best-RowdenJ17, DBLP:journals/corr/abs-1904-01740}, viewpoint \cite{DBLP:journals/corr/abs-1812-10902}, and illumination \cite{OTOOLE2018794}), demographics \cite{DBLP:conf/eccv/DasDB18, BTAS_terhoerst, DBLP:conf/biosig/OzbulakAE16}, and social traits \cite{DBLP:journals/cogsci/PardeHCSO19}.
% DBLP:conf/fusion/TerhorstHKDKK19, terhrst2020serfiq, Terhorst2020FaceQE
% 4 - describe the general problem are or the current research focus of the field
However, for many applications, the users do not permit to have access to this information. 
Thus, the stored data should be exclusively used for recognition purposes \cite{DBLP:conf/icb/MirjaliliR17}, and extracting such information without a person's consent is considered a violation of their privacy \cite{Kindt2013}.
This problem is known as soft-biometric privacy \cite{DBLP:conf/icb/MirjaliliR17} and solutions are either build on image- \cite{DBLP:conf/eccv/OthmanR14, DBLP:journals/access/MirjaliliRR19, mirjalili2020privacynet} or template-level  \cite{ICB_Terhoerst_IVE, DBLP:journals/corr/abs-2002-09181, DBLP:journals/access/TerhorstRDRBKSK20, FG_privacy}.
% Terhörst2019, DBLP:journals/corr/abs-2002-09181
 
% 5 - provide a transition between the general problem area and the literature review
% 6 - provide a brief overview of key research projects in this area
% 7 - describe research gap

Since the knowledge about encoded attributes in face template is required to develop more advanced bias-mitigating solutions \cite{gong2019debface,Liang_2019_CVPR, DBLP:conf/iwbf/TerhorstTDKK20, DBLP:journals/corr/abs-2002-03592,DBLP:conf/cvpr/00010S0C19} and more comprehensive privacy-enhancing technologies, in this work we investigate the predictability of 113 attributes from face templates at different difficulty-levels.
We jointly trained a massive attribute classifier (MAC) with a high number of attributes to take advantage of a shared feature space.
The MAC is modified such that it is able to accurately state its prediction reliability \cite{BTAS_terhoerst}.
This allows us to make predictions at two reliability levels and thus, to derive more fine-grained statements about the predictability of attributes in face templates.
The experiments were conducted on two publicly available databases, CelebA \cite{liu2015faceattributes} and LFW \cite{LFWTech}, and on two popular face embeddings, FaceNet \cite{DBLP:journals/corr/SchroffKP15} and ArcFace \cite{Deng_2019_CVPR}.
To derive understandable statements about the stored attribute information, we categorized each attribute into on of three predictability classes: easily-predictable, predictable, and hardly-predictable.
The results shows that 39 attributes are assigned to the easily-predictable class and 74 of the 113 investigated attributes are at least predictable.
Despite that face templates are learned to be robust to non-permanent factors, the results demonstrate that especially these attributes are easily-predictable.
This includes information about age, hairstyles, haircolors, beards, and accessories, such as makeup, lipstick, and glasses.
%Future works on face recognition might focus on suppressing these non-permanent attributes to prevent biased decisions and preserve the privacy of the users.

%introduce predictability classes
%we can only make statements if a attribute is predictable not the other way around

%%%%%%%%% Related work
\section{Related work}
\label{sec:RelatedWork}
%%% shows how your contribution extends state of the art
%%% where is the gap to be filled
%%% also indicates how familiar you are with the topic
%%% mention weak points in related work

The development of deep neural network representations for faces led to strong performance boosts for face recognition \cite{FRVT2018}.
However, since these representations are derived from black-box models, it is not clear which kind of information is stored in these representations.

In 2017, Parde et al. \cite{DBLP:conf/fgr/PardeCHCSCO17} investigated face representations in terms of head position and source of the image.
The results demonstrated that the investigated representations contain accurate information of the yaw and pitch of a face and about whether the input-face origins from a still image or a video frame.
They suggest that image-quality information might available in these features as well.
% image quality
This hypothesis was proofed to be correct \cite{terhrst2020serfiq, DBLP:journals/corr/Best-RowdenJ17, DBLP:journals/corr/abs-1904-01740}. 
In \cite{terhrst2020serfiq, DBLP:journals/corr/Best-RowdenJ17, DBLP:journals/corr/abs-1904-01740}, face image quality was successfully predicted based on face embeddings.
%In \cite{terhrst2020serfiq}, Terh\"orst et al. demonstrated that the face quality correlates with the stability of its face embeddings and that the qualities reflect the same biases then the embeddings \cite{Terhorst2020FaceQE} bias.

%2019
In \cite{DBLP:journals/cogsci/PardeHCSO19}, Parde et al. analysed if face representations retain information in faces that supports social-trait inferences.
In their experiments, they investigated 11 social traits such as talkative, assertive, shy, quiet, warm, artistic, efficient, careless, impulsive, anxious, and lazy. 
They trained linear classifiers to predict these human-assigned social trait profiles and demonstrated that these traits can be determined from face embeddings to a high degree.
The best-predicted traits were impulsive, warm, and anxious.

Hill et al. \cite{DBLP:journals/corr/abs-1812-10902} analysed the representations of caricature faces.
They examined the organization of viewpoint (0°, 20°, 30°, 45°, 60°), illumination (ambient vs spotlight), gender (male vs female), and identity in the embedding space.
Their results showed that the utilized face recognition model creates a highly organized, hierarchical, similarity structure in which information about face identity and imaging characteristics coexist.
These results were summarized by O'Toole et al. \cite{OTOOLE2018794}.
They reviewed what properties are known about the face space and ground them in the context of previous-generation face recognition algorithms.

In \cite{DBLP:conf/icb/ZhongSL16, DBLP:conf/icip/ZhongSL16} Zhong et al. demonstrated that the use of various mid-level representations from face recognition networks leads to highly accurate facial attribute estimation performances.
This indicates that also high-level representations, such as face recognition templates, might contain a significant amount of facial attribute information.
% demographic information
In \cite{DBLP:conf/eccv/DasDB18, BTAS_terhoerst, DBLP:conf/fusion/BoutrosDTKK19, DBLP:conf/biosig/OzbulakAE16}, it is shown that demographic attributes such as gender, age, and race can be derived from face templates.
% \cite{DBLP:conf/eccv/DasDB18, BTAS_terhoerst, DBLP:conf/fusion/TerhorstHKDKK19, DBLP:conf/biosig/OzbulakAE16}

%Gender age race on template level 
%age gender race --> add papers that use face templates (BTAS, FUSION, bias antiza
%
%age gender race + bias mitigation based on face template Das et al. \cite{DBLP:conf/eccv/DasDB18}
%age and gender with reliability statements about the prediction confidence Terh\"orst et al. \cite{BTAS_terhoerst, DBLP:conf/fusion/TerhorstHKDKK19}
%
%fine-tuned face recognition model \cite{DBLP:conf/biosig/OzbulakAE16} for age and gender estimation

So far, previous work showed that head pose, image characteristics (such as quality, source of the image, viewpoint, illumination), demographic attributes (gender, age, race), and social traits (e.g. impulsive, warm, and anxious) can be found in face templates.

In contrast to previous work that investigated only specific characteristics, in this work we analyse a wide range of attributes (up to 113) in face representations.
Moreover, we analyse the predictability of these attributes under different levels of prediction reliabilities. 
This allows us to state more generally which attributes are encoded in face templates.

\section{Investigation methodology}
%%% communicate about your procedure/method/approach
%%% write for people with less knowledge than you about this topic
%%% everyone who reads it should be able to repeat the experiments, understand and accept the work

This work aims at analysing the set of soft-biometric information that is stored in face templates.
To do so we train a classifier to jointly predict these attributes.
If the classifier can successfully predict these, we conclude that these attributes are stored in the face templates.
However, this only allows us to answer the question of what information is embedded. 
A statement about what information is not included is not possible, because the reverse conclusion is not necessarily logical. 
If an estimator is not able to learn the pattern of an attribute, it does not imply that the pattern does not exist. 
The classifier might just not be able to deal with the complexity of the attribute pattern or the data variability and representation might be low.

To answer the research question of this work, the following three subsections explain the different steps of the investigation methodology.
In Section \ref{sec:MAC}, we will first explain the classifier training procedure that allows a joint prediction of a large number of attributes.
Learning these attributes in a multi-task learning approach will enhance the performance, since many attributes share similar features.
In Section \ref{sec:Reliability}, we explain how this classifier can accurately state its predictions confidence.
This prediction confidence determines the quality of a prediction and enables us to derive predictability classes in Section \ref{sec:PredictabilityClasses}.
These predictability classes allow us to generalize our findings into easily understandable statements.

%
% allows us to determine capturing conditions and 
%
%repeat goal
%1. classifier
%2. reliability statements to determine capturing conditions
%3. predictability classes to generalize conclusions

\subsection{Massive attribute classifier (MAC)}
\label{sec:MAC}
To investigate what attribute-information is stored in face templates, we train a classifier model to predict multiple attributes.
If the classifier can correctly predict these attributes given face templates, we can draw conclusions about what attributes are encoded in the investigated representation.

Therefore, we trained a neural network model to jointly predict multiple attributes given face templates of the training set.
Due to the large number of predicted attributes, we refer to this model as the massive attribute classifier (MAC).
To find an optimal network structure for our MAC, we evaluated multiple models with various number of dense layers and layer sizes.
To be precise, we evaluated random network structures with 1-3 initial layers and 1-3 branch layers that connects the last initial layer with the the softmax layers of each attribute.
For each layer a size of 128, 256, and 512 was evaluated.
We choose the structure with the most stable results as the layout of our MAC.
However, despite the large variations in the investigated network structures, we observed that, in most cases, the predicted performance per attribute only varies within a range of 1-2\%.

The chosen MAC-network consists of two initial layers, the input layer of size $n_{in}$ and the second dense layer of size 512.
Here, $n_{in}$ refers to the size of the utilized face embedding.
Starting from the second layer, each attribute $a$ has an own branch consisting of two additional layers of size 512 and $n_{out}^{(a)}$,
where $n_{out}^{(a)}$ refers to the number of classes per attribute.
Each layer has a ReLU activation, except for the output-layers, which have softmax activations.
Moreover, Batch-Normalization \cite{DBLP:conf/icml/IoffeS15} and dropout \cite{Srivastava:2014:DSW:2627435.2670313} with a dropout-probability of $p_{drop} = 0.5$ is applied to every layer.
The dropout allows to generalize the performance, but also enables us to derive reliability statements about the predictions (described in Section \ref{sec:Reliability}).
The training of the MAC was done in a multi-task learning fashion by applying a categorical cross-entropy loss for each attribute branch and use an equal weighting between each of these attribute-related losses.
For the training, an Adam optimizer \cite{DBLP:journals/corr/KingmaB14} was used with $e=200$ epochs, an initial learning rate $\alpha = 10^{-3}$, and a learning-rate decay of $\beta = \alpha/e$.
These parameter choices are guided by \cite{BTAS_terhoerst}.
The batch size $b$ was chosen according to the amount of data available, $b=1024$ for CelebA and $b=16$ for LFW.

\subsection{Reliability statements}
\label{sec:Reliability}
To derive statements about the predictability of an attribute in a face template, we use prediction reliabilities to simulate close-to-optimal classifier circumstances.
Therefore, we follow the methodology in \cite{BTAS_terhoerst, DBLP:conf/fusion/TerhorstHKDKK19} to enable our MAC to state its prediction confidence (reliability).
Following this approach, we trained the MAC with dropout.
To derive a reliability statement additionally to an attribute prediction, $m=100$ stochastic forward passes are performed.
In each forward pass, a different dropout-pattern is applied, resulting in $m$ different softmax outputs $v_i^{(a)}$ for each attribute $a$.
Given the outputs of the $m$ stochastic forward passes of the predicted class $\hat{c}$ denoted as $x^{(a)}=v_{i,\hat{c}}^{(a)}$, the reliability measure is given as 
\begin{align*}
rel(x^{(a)}) =   \dfrac{1-\alpha}{m} \sum_{i=1}^m x_i^{(a)} - \dfrac{\alpha}{m^2} \sum_{i=1}^m \sum_{j=1}^m |x_i^{(a)} - x_j^{(a)}|,
\end{align*}
with $\alpha=0.5$, following the recommendation in \cite{BTAS_terhoerst}.
The first part of the equation is a measure of centrality and utilizes the probability interpretation of the softmax output.
A higher value can be interpreted as a high probability that the prediction is correct.
The second part of the equation is the measure of dispersion and quantifies the agreement of the stochastic outputs $x$.
In \cite{BTAS_terhoerst}, this was shown to be an accurate reliability measure.
%and \cite{DBLP:conf/fusion/TerhorstHKDKK19}

We use this reliability measure to simulate more idealistic circumstances.
For each attribute, we calculate the prediction and corresponding reliability of each instance.
Then we take the predictions of 100\% and 50\% of the highest reliabilities to evaluate the performance.
This performance refers to the ratio of considered predictions (RCP) of 100\% and 50\%.
The performance at 100\% RCP refers to the general performance of the whole dataset.
The performance at 50\% RCP refers to the performance on the predictions with 50\% of the highest reliabilities.
Consequently, this refers to the performance based on the prediction on which the MAC is most confident about.
The unconsidered 50\% of the predictions might contain factors of variances (such as blur, non-frontal head poses) that lead to unstable, and thus inaccurate, attribute estimates.

\subsection{Predictability classes}
\label{sec:PredictabilityClasses}

To derive more understandable statements about which attribute information is stored in a face template, we categorize each attribute into one of three predictability classes: 
\begin{itemize}
\item \textbf{Easily-predictable (++):} an attribute is categorized as easily-predictable if, and only if, the balanced accuracy at 100\% RCP is above 90\%. This means that highly \textit{accurate predictions are possible even under non-ideal circumstances} such as bad illuminations and non-frontal head poses.
\item \textbf{Predictable (+):} an attribute is categorized as predictable if, and only if, the balanced accuracy at 100\% RCP is under 90\%, but the balanced accuracy at 50\% RCP is above 90\%. This indicates that highly \textit{accurate predictions are possible under close-to-optimal conditions}, since it only takes into account 50\% of the most confident MAC predictions.
\item \textbf{Hardly-predictable (0):} an attribute is categorized as hardly-predictable if the balanced accuracy is below 90\% at both, 100\% and 50\% RCP. Even \textit{under close-to-optimal circumstances, the MAC is not able to reach high accuracies}. Consequently, the attribute patterns might be too complex for the MAC to handle or it does not exist a meaningful pattern for this attribute.
\end{itemize}
While the first two categorizes (Easily-predictable and Predictable) allow making confident statements about the amount of attribute information in face templates, the same does not apply for the third category (Hardly-predictable).
The last category only states that the classifier is not able to accurately learn the pattern, but this might be due to several reasons: (1) the pattern does not exist, (2) the pattern does exist, but it is too complex for the model to learn, or (3) the pattern does exists but the amount of data and its representation is not appropriate for the classifier to learn.
Consequently, for the third case, we can not determine if the attribute pattern exists.

%This allows to quickly 
%
%Introduction of the predictability classes
%
%easily-predictable - even under difficult circumstances (bad illumination, non-frontal head pose,...) it achieves high accuracies.
%
%predictable - under close-optimal circumstances it reaches high accuracies
%
%hardly-predictable - under close-optimal circumstances it does not reach high accuracies
%but the attribute patterns might be to complex for our MAC to handle.

%%%%%%%%% Experimental setup
\section{Experimental setup} 
\label{sec:ExperimentalSetup}
%%% explain in detail how you did the experiments
%%% they must be reproducable, understandable, accepted by the reader

% 1 - explain the database used and the benefits, why this one?
\subsection{Databases}

%% Sample images
\begin{figure*}[t]
\centering
% CelebA
\includegraphics[width=0.0955\textwidth]{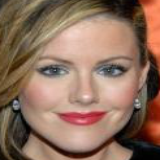}
	\hfill
	\includegraphics[width=0.0955\textwidth]{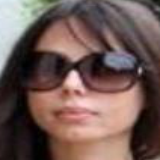}
	\hfill 
	\includegraphics[width=0.0955\textwidth]{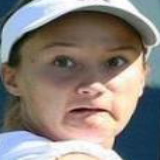}
	\hfill 
	\includegraphics[width=0.0955\textwidth]{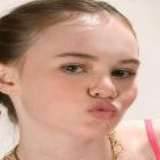}
	\hfill 
	\includegraphics[width=0.0955\textwidth]{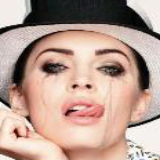}
	\hfill 
	\includegraphics[width=0.0955\textwidth]{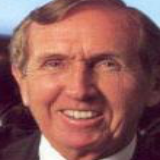}
	\hfill 
	\includegraphics[width=0.0955\textwidth]{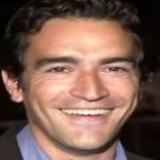}
	\hfill 
	\includegraphics[width=0.0955\textwidth]{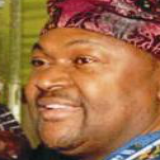}
	\hfill 
	\includegraphics[width=0.0955\textwidth]{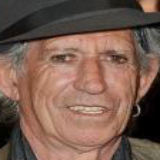}
	\hfill 
	\includegraphics[width=0.0955\textwidth]{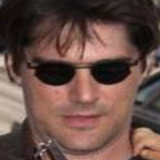}
	\hfill  
% LFW
\includegraphics[width=0.0955\textwidth]{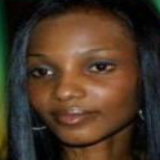}
	\hfill
	\includegraphics[width=0.0955\textwidth]{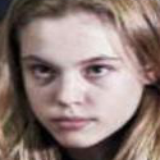}
	\hfill 
	\includegraphics[width=0.0955\textwidth]{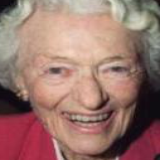}
	\hfill 
	\includegraphics[width=0.0955\textwidth]{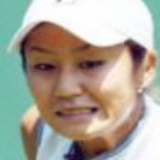}
	\hfill 
	\includegraphics[width=0.0955\textwidth]{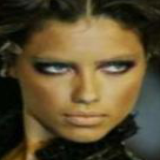}
	\hfill 
	\includegraphics[width=0.0955\textwidth]{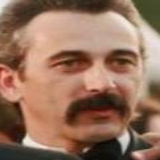}
	\hfill 
	\includegraphics[width=0.0955\textwidth]{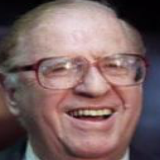}
	\hfill 
	\includegraphics[width=0.0955\textwidth]{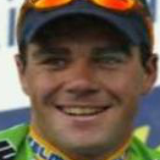}
	\hfill 
	\includegraphics[width=0.0955\textwidth]{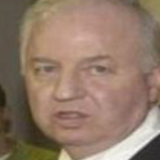}
	\hfill 
	\includegraphics[width=0.0955\textwidth]{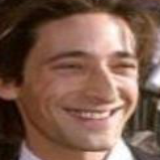}
	\hfill    
\vspace{-4mm}
\caption{Sample images from CelebA (top row) and LFW (bottom row)} \vspace{-3mm}
\label{fig:SamplesImages}
\end{figure*}

For the analysis of the face space, we chose the Labeled Faces in the Wild (LFW) \cite{LFWTech} and the CelebFaces Attributes (CelebA) \cite{liu2015faceattributes} datasets because of their large and rich attribute annotations.
The large number of different soft-biometric labels allows to deeply investigate which of these attributes are encoded in face templates.
Figure \ref{fig:SamplesImages} shows sample images from both datasets.
The CelebA dataset \cite{liu2015faceattributes} is a large-scale dataset with more than 200k images of over 10k celebrities.
It covers large variations in pose and background.
Moreover, each image is labelled with 40 binary attributes.
LFW \cite{LFWTech} contains over 13k images from over 5k individuals and exhibits variability in pose, lighting, focus, resolution, facial expression, age, gender, race, accessories, make-up, occlusions, background, and photographic quality.
The face images are 250x250 pixels and mostly in color.
Each image is annotated with up to 73 attributes.

The attribute labels of both databases \cite{LFWTech, liu2015faceattributes} cover a wide range of characteristics such as the person's demographics, skin, hair, beard, face geometry, periocular area, mouth, nose, accessories, and environment.

\subsection{Cleaning attribute labels}
\label{sec:AttributeCleaning}
In contrast to CelebA, where the attribute labels are of binary nature, in LFW, the labels come from the prediction probabilities of a binary classifier \cite{LFWTech}.
Each label value measures the degree of the attribute and thus, are continuous \cite{DBLP:conf/iccv/KumarBBN09, DBLP:journals/pami/KumarBBN11}.
E.g. for the attribute male, a higher label score indicates that the person appears more masculine than a person with a lower label score.
Consequently, the top rank images for an attribute represent the label true, while the lowest rank images indicate the label false.
A value around zero means that the corresponding attribute has little meaning on this image.

To make sure that our MAC performs well when training on LFW, we manually converted the continuous attribute labels to binary labels.
Therefore, we assigned an upper and lower score threshold for each attribute.
Images with a score over the upper threshold are assigned as true, images with a score under the lower threshold are assigned as false, images with scores within the range are assigned as undefined.
The upper and lower thresholds for one attribute are manually determined by moving potential thresholds away from zero.
At each potential threshold, ten images with the closest attribute scores are investigated.
Here, the original LFW labels of the images are manually investigated for correctness.
If only eight or fewer attributes are investigated as correct, the potential threshold is further moved away from the starting point and the procedure is repeated.
If a potential threshold returns images with 9 or more correct labels, it is chosen as the limit.
%The correctness of the labels is evaluated manually.
Repeating this over all attributes will result in a lower and an upper threshold for each of these attributes.
By binaryzing the scores with these upper and lower thresholds, we ensure an error-minimizing data basis of the MAC.
This allows us to train and test on meaningful and correctly labelled data.

Please note that the label-cleaning process reduces the amount of used labels by 51,7\% that might induce a bias in our evaluation. 
To avoid biased conclusions that might result from this process, we evaluate on another binary labelled database.
After the label-cleaning, we found 15 attribute labels of either a low number of positively and negatively labelled samples ($<$100). 
These are listed in Table \ref{tab:TrainTestLFW_selected} with the number of positively and negatively labelled samples in the test and training set.
%Our manual sample-filtering approach leads to either very low number of positively or negatively labelled samples.
We will mark these attributes (in grey) in the following investigations to consider their low expressiveness during the face analysis.

%classifiers trained on single feature-region combinations
%trained with SVM classifiers

\begin{table}[]
\setlength{\tabcolsep}{5pt}
\renewcommand{\arraystretch}{1.0}
\centering
\caption{Train/test sample distribution on LFW for \textit{selected attributes} that are found insufficient for a meaningful attribute analysis \textit{after label-cleaning}. Pos and Neg refers to the number of positively and negatively labelled samples for the train and test set. The listed 15 attributes are found to be insignificant for the analysis due to a low number of samples in either the positive or negative class.}
\label{tab:TrainTestLFW_selected}
\begin{tabular}{lrrrr}
\Xhline{2\arrayrulewidth}
                             & \multicolumn{2}{c}{Train} & \multicolumn{2}{c}{Test} \\
                             \cmidrule(rl){2-3} \cmidrule(rl){4-5}
Attribute                    & Pos     & Neg     & Pos    & Neg   \\
\hline
Color Photo                 & 8806        & 29          & 3772        & 24         \\
Mouth Slightly Open        & 674         & 109         & 315         & 57         \\
Round Face                  & 9           & 588         & 3           & 250        \\
Goatee                       & 20          & 3346        & 10          & 1557       \\
Baby                         & 23          & 9137        & 15          & 3913       \\
Bangs                        & 89          & 5238        & 44          & 2080       \\
Bald                         & 114         & 4413        & 47          & 1953       \\
Big Lips                    & 101         & 751         & 48          & 318        \\
Sunglasses                   & 74          & 8583        & 50          & 3631       \\
Partially Visible F. & 124         & 1501        & 55          & 601        \\
Mouth Wide Open            & 107         & 6593        & 56          & 2925       \\
Double Chin                 & 154         & 172         & 57          & 136        \\
Harsh Lighting              & 113         & 914         & 62          & 487        \\
Outdoor                      & 173         & 510         & 63          & 243        \\
Teeth Not Visible          & 125         & 2209        & 66          & 1089       \\
\Xhline{2\arrayrulewidth}
\end{tabular}
\end{table}

% 2 - explain the evaluation metric
\subsection{Evaluation metrics}
In this work we derive what information is contained in the face templates based on prediction accuracies.
In machine learning, accuracy is defined by the ratio of the number of correct predictions to the total number of predictions \cite{murphy2013machine}.
To be robust to attribute-imbalances, we report the prediction performance in terms of balanced accuracy.
This refers to the standard accuracy with class-balanced sample weights \cite{10.5555/2815672}.

The train/test data is defined by dividing the databases in a 70\%/30\% subject-exclusive split. 
To analyse the prediction performance of an estimator under more ideal circumstances, we chose a classifier for the attribute prediction task that is additionally able to accurately state its prediction confidence.
For each face template, this classifier predicts the associated attributes and their prediction reliabilities.
To get the prediction performance under more ideal circumstances, for each attribute, only the predictions with 50\% of the highest reliabilities are considered for the balanced accuracy.
This balanced accuracy refers to a ratio of considered predictions (RCP) of 50\%.
Since this relates to the MAC prediction confidence, the balanced accuracy should be higher at lower RCP-levels.

\subsection{Face template extraction}
\vspace{-1mm}
In this work, we utilize two widely-used face recognition models, FaceNet \cite{DBLP:journals/corr/SchroffKP15} and ArcFace \cite{Deng_2019_CVPR}.
We use pre-trained models trained on the MS1M database \cite{DBLP:journals/corr/GuoZHHG16} for both networks, FaceNet\footnote{\url{https://github.com/davidsandberg/facenet}} and ArcFace\footnote{\url{https://github.com/deepinsight/insightface}}.
To get the face template for a given face image, the image has to be aligned, scaled, and cropped. 
For FaceNet, the preprocessing is done as described in \cite{Kazemi2014OneMF}.
For ArcFace, we follow the preprocessing as described in \cite{DBLP:journals/corr/abs-1812-01936}.
The preprocessed image is passed to a face recognition model to extract the embeddings.
The output size is 128 for FaceNet and 512 for ArcFace.

% 3 - what was investigated and why?
\subsection{Investigations}
\vspace{-1mm}
This works aims at understanding what kind of soft-biometric information is stored in face templates. Therefore, our investigations are divided into three parts:
\begin{enumerate}
\item We validate the attributes labels of both datasets by studying the correlations between the attributes.
%The correlation of the attribute labels of both databases is analysed to obtain the validity of the following attribute predictions.
\item We analysing what attributes are contained in face representations by investigating the attribute prediction performances on both datasets and face embeddings.
To get a more complete perspective on the problem, the prediction performances on different confidence-levels of the classifier are investigated.
%The attribute prediction performances on both datasets and face embeddings are investigated. 
%This aims to answer which attributes are contained in such face representations.
%To get a more complete perspective on the problem, the prediction performances on different confidence-levels of the classifier are investigated.
\item We obtain an overview of which kind of information is encoded in face templates by categorizing each attribute into one of three predictability classes based on their two-level prediction performances.
%
%The two-level prediction performances are categorized into three predictability classes to obtain an overview of which kind of information is encoded in face templates.
\end{enumerate}

%%%%%%%%% Results
\section{Results}
\label{sec:Results}
\vspace{-1mm}
%This section is divded into three subsections, each focusing on one investigation point.
%Section \ref{sec:CorrelationAnalysis} analyses attribute-label correlations, Section \ref{Sec:Attribute-analyses} investigates predictability of these attributes, and Section \ref{sec:Summary} provides are summary of the findings.

This section is divided into three subsections, each focusing on one investigation point:
(1) analysis of the attribute correlation, (2) investigation of the attribute predictability, and (3) summarize findings.

\subsection{Attribute-correlation analysis}
\label{sec:CorrelationAnalysis}
\vspace{-1mm}

%% face quality distributions
\begin{figure*}[t]
\centering
  \subfloat[CelebA \label{fig:correlationCelebA}]{%
       \includegraphics[width=0.45\textwidth]{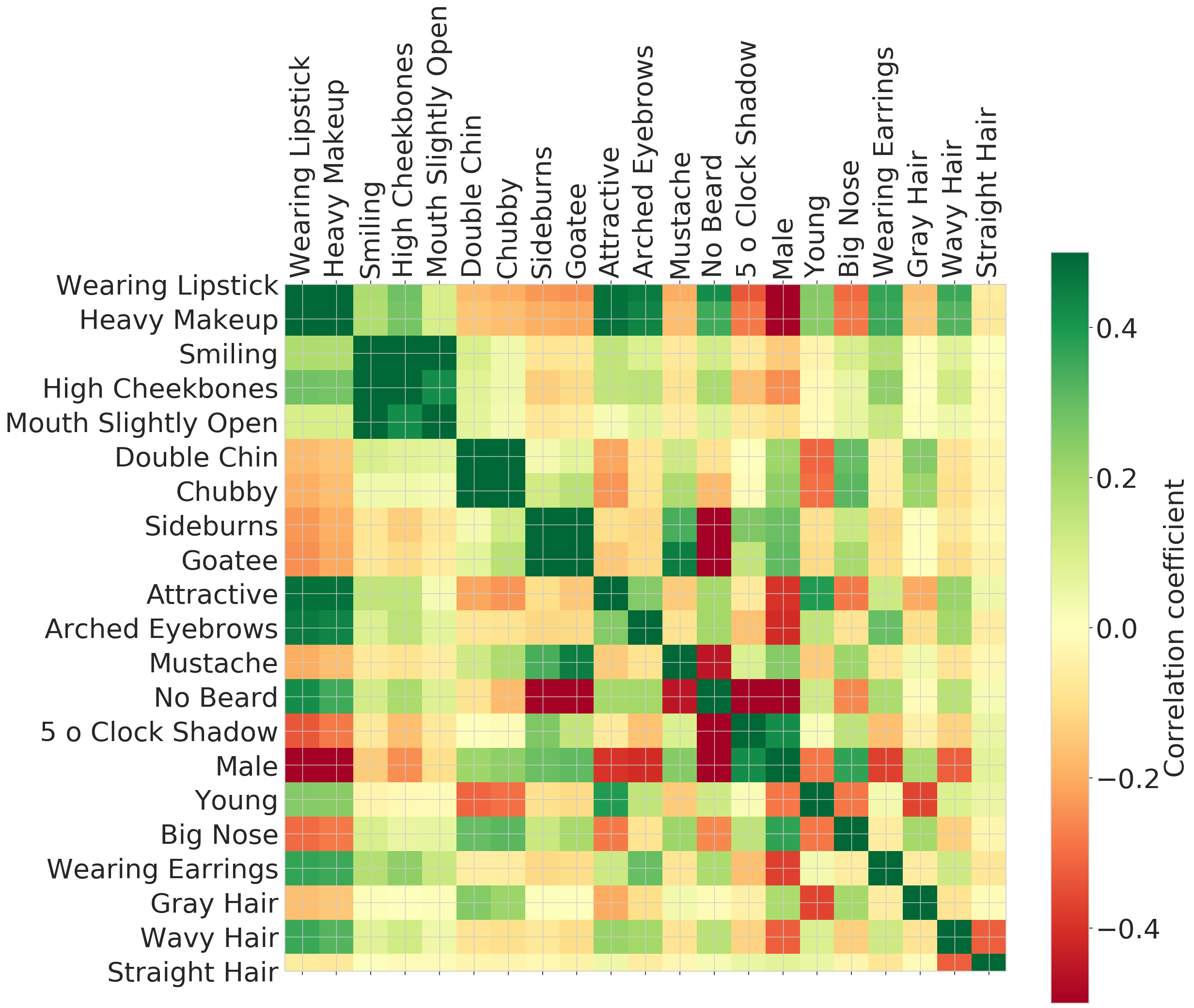}} 
	\hfill       
  \subfloat[LFW \label{fig:correlationLFW}]{%
       \includegraphics[width=0.45\textwidth]{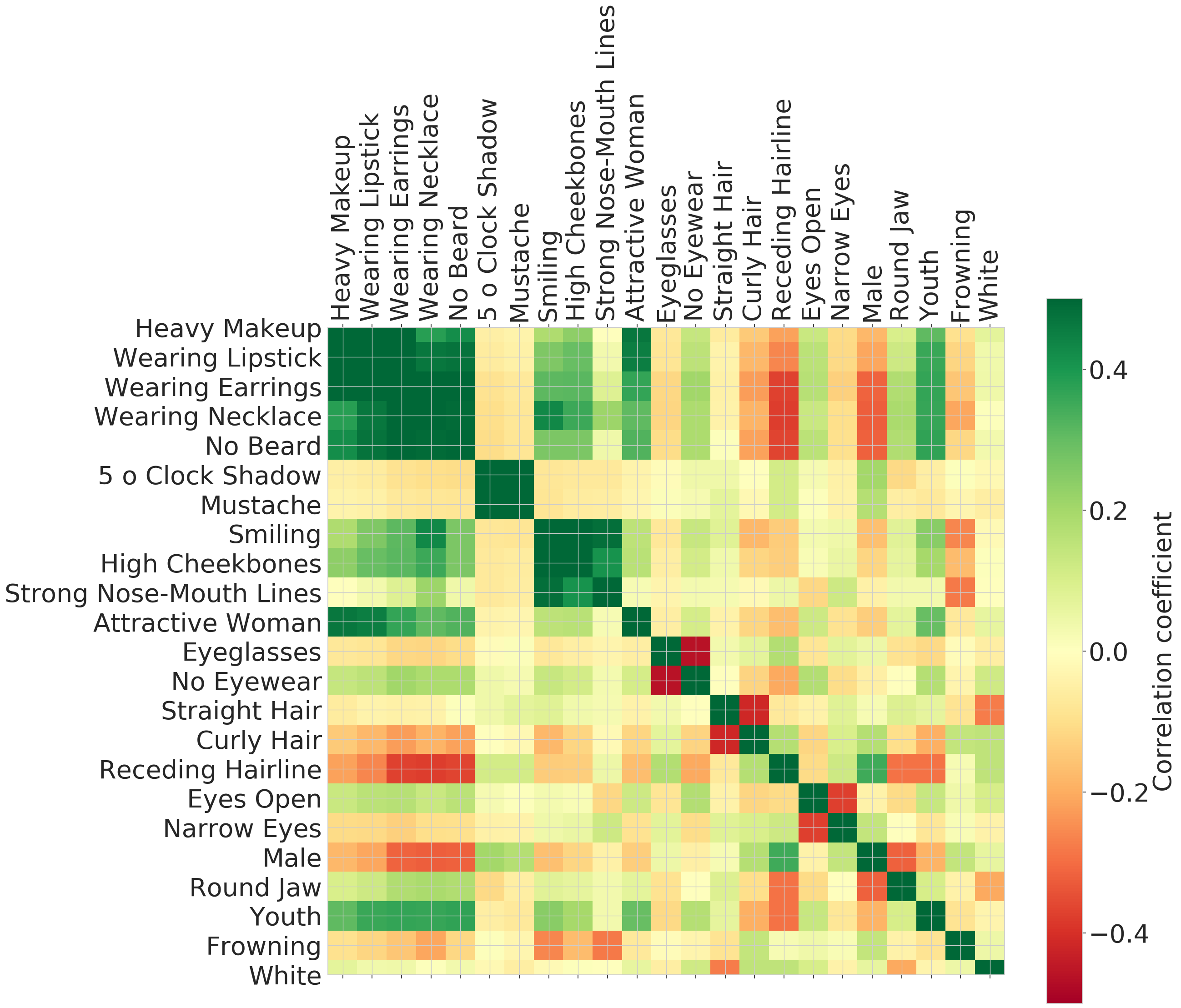}} 
\caption{Label-correlation for CelebA and LFW. The attributes are chosen to show the 15 most positive and negative pairwise correlations. The attribute-correlation for LFW is shown after the label-cleaning process. Green indicate positive correlations, while red indicate a negative correlation. The correlation is based on the Pearson coefficient.} \vspace{-2mm}
\label{fig:Correlation matrix}
\end{figure*}

To understand the quality of the labels and potential biases in the attribute labels, Figure \ref{fig:Correlation matrix} shows a selection of attribute-label correlations.
The attributes are chosen to show the 15 most positive and negative pairwise correlations.
For CelebA, the correlation in Figure \ref{fig:correlationCelebA} shows that the large majority of male faces in the database do not wear lipstick, earrings, and makeup.
These attributes mostly belong to female faces.
Moreover, it shows some biases in the database labels.
The majority of male faces have a beard.
If a face is labelled as \textit{attractive}, it belongs to a young female face most likely wearing accessories and makeup.
However, this figure also approves the quality of some labels.
E.g. \textit{No Beard} negatively correlates with all kinds of beards such as \textit{Sideburns}, \textit{Goatee}, and \textit{Mustache}.

Figure \ref{fig:correlationLFW} shows the attribute correlation for LFW.
It shows that the attributes \textit{Heavy Makeup}, \textit{Wearing Lipsticks}, \textit{Wearing Earrings}, and \textit{Wearing Necklace} belongs together with \textit{Youth} and \textit{Attractive Woman}, \textit{Smiling}, and \textit{High Cheekbones}.
Moreover, this set of attributes does not correlate with a \textit{Receding Hairline} and \textit{Male}.
Nevertheless, it also approves the quality of other labels such as \textit{No Eyewear} (negatively correlates with \textit{Eyeglasses}) and \textit{Curly Hairs} (negatively correlates with \textit{Straight Hair}).

\subsection{Attribute-analysis of the face space}
\label{Sec:Attribute-analyses}

\begin{table}[]
\setlength{\tabcolsep}{3pt}
\renewcommand{\arraystretch}{1.0}
\centering
\caption{Prediction performance on CelebA: the performance is based on FaceNet (FN) and ArcFace (AF) embeddings and is reported in terms of balanced accuracies at two difficulty scenarios: 100\% RCP (hard) and 50\% RCP (easy). $^{++}$,$^{+}$, and $^{0}$ state the assigned predictability class.}
\label{tab:PerformanceCelebA}
\begin{tabular}{llrrrr}
\Xhline{2\arrayrulewidth}
 & & \multicolumn{2}{c}{100\% RCP} & \multicolumn{2}{c}{50\% RCP} \\
 \cmidrule(rl){3-4} \cmidrule(rl){5-6}
 & Attribute & \multicolumn{1}{c}{FN}       & \multicolumn{1}{c}{AF}               & \multicolumn{1}{c}{FN} & \multicolumn{1}{c}{AF}         \\
\hline
\parbox[t]{2mm}{\multirow{2}{*}{\rotatebox[origin=c]{90}{Demo}}}  & Male$^{++}$                  & 98.9\%  & 98.4\%  & 99.9\%  & 99.9\%  \\
              & Young$^{+}$                & 85.5\%  & 83.6\%  & 96.4\%  & 94.5\%  \\ \rule{0pt}{3ex}\noindent
\parbox[t]{2mm}{\multirow{2}{*}{\rotatebox[origin=c]{90}{Skin}}}        & Pale Skin$^{0}$            & 76.0\%  & 71.9\%  & 87.1\%  & 83.0\%  \\
              & Rosy Cheeks$^{+}$          & 83.4\%  & 78.2\%  & 96.3\%  & 81.7\%  \\ \rule{0pt}{3ex}\noindent
\parbox[t]{2mm}{\multirow{6}{*}{\rotatebox[origin=c]{90}{Hairstyle}}}     & Bald$^{++}$                  & 95.7\%  & 94.0\%  & 100.0\% & 100.0\% \\
              & Bangs$^{++}$                 & 91.7\%  & 89.3\%  & 99.4\%  & 98.3\%  \\
              & Receding Hairline$^{+}$    & 85.4\%  & 82.5\%  & 96.4\%  & 94.2\%  \\
              & Sideburns$^{++}$             & 92.8\%  & 92.1\%  & 90.0\%  & 99.7\%  \\
              & Straight Hair$^{0}$        & 68.6\%  & 70.7\%  & 79.9\%  & 82.0\%  \\
              & Wavy Hair$^{0}$            & 74.4\%  & 76.6\%  & 86.4\%  & 89.4\%  \\ \rule{0pt}{3ex}\noindent
\parbox[t]{2mm}{\multirow{4}{*}{\rotatebox[origin=c]{90}{Haircolor}}}      & Black Hair$^{+}$           & 83.7\%  & 81.5\%  & 96.6\%  & 94.3\%  \\
              & Blond Hair$^{++}$           & 91.9\%  & 90.1\%  & 99.3\%  & 98.3\%  \\
              & Brown Hair$^{+}$           & 76.5\%  & 75.9\%  & 90.1\%  & 88.3\%  \\
              & Gray Hair$^{++}$            & 93.0\%  & 91.1\%  & 99.6\%  & 98.8\%  \\ \rule{0pt}{3ex}\noindent
\parbox[t]{2mm}{\multirow{4}{*}{\rotatebox[origin=c]{90}{Beard}}}          & 5 o Clock Shadow$^{+}$   & 86.9\%  & 85.8\%  & 99.6\%  & 99.0\%  \\
              & Goatee$^{++}$                & 93.4\%  & 91.8\%  & 97.2\%  & 98.9\%  \\
              & Mustache$^{++}$              & 92.2\%  & 89.7\%  & 100.0\% & 98.8\%  \\
              & No Beard$^{++}$             & 92.1\%  & 90.8\%  & 99.4\%  & 99.0\%  \\  \rule{0pt}{3ex}\noindent
\parbox[t]{2mm}{\multirow{4}{*}{\rotatebox[origin=c]{90}{Face Geo.}}}  & Chubby$^{+}$                & 86.5\%  & 83.1\%  & 96.5\%  & 95.4\%  \\
              & Double Chin$^{+}$          & 86.6\%  & 82.9\%  & 96.9\%  & 95.4\%  \\
              & High Cheekb.$^{+}$      & 78.5\%  & 72.2\%  & 91.6\%  & 82.6\%  \\
              & Oval Face$^{0}$            & 63.4\%  & 61.9\%  & 70.8\%  & 68.1\%  \\ \rule{0pt}{3ex}\noindent
\parbox[t]{2mm}{\multirow{4}{*}{\rotatebox[origin=c]{90}{Periocular}}}      & Arched Eyebrows$^{+}$      & 79.8\%  & 77.0\%  & 93.3\%  & 89.5\%  \\
              & Bags Under Eyes$^{0}$     & 72.1\%  & 70.7\%  & 80.6\%  & 80.7\%  \\
              & Bushy Eyebrows$^{+}$       & 83.4\%  & 78.5\%  & 95.9\%  & 91.9\%  \\
              & Narrow Eyes$^{0}$          & 66.5\%  & 60.7\%  & 75.4\%  & 66.7\%  \\ \rule{0pt}{3ex}\noindent
\parbox[t]{2mm}{\multirow{3}{*}{\rotatebox[origin=c]{90}{Mouth}}}          & Big Lips$^{0}$             & 74.6\%  & 68.8\%  & 86.4\%  & 78.7\%  \\
              & Mouth Slightly Open$^{0}$ & 74.5\%  & 67.5\%  & 86.5\%  & 76.5\%  \\
              & Smiling$^{+}$               & 80.1\%  & 71.7\%  & 92.9\%  & 82.1\%  \\ \rule{0pt}{3ex}\noindent
\parbox[t]{2mm}{\multirow{2}{*}{\rotatebox[origin=c]{90}{Nose}}}           & Pointy Nose$^{0}$          & 71.7\%  & 69.3\%  & 83.1\%  & 78.9\%  \\
              & Big Nose$^{0}$             & 77.4\%  & 75.8\%  & 88.1\%  & 87.1\%  \\ \rule{0pt}{3ex}\noindent
\parbox[t]{2mm}{\multirow{7}{*}{\rotatebox[origin=c]{90}{Accessories}}}    & Eyeglasses$^{++}$            & 97.3\%  & 90.6\%  & 99.8\%  & 98.7\%  \\
              & Heavy Makeup$^{++}$         & 90.1\%  & 88.7\%  & 99.2\%  & 98.5\%  \\
              & Wearing Earrings$^{+}$     & 79.2\%  & 77.0\%  & 94.8\%  & 91.6\%  \\
              & Wearing Hat$^{++}$          & 95.4\%  & 92.8\%  & 99.4\%  & 99.0\%  \\
              & Wearing Lipstick$^{++}$     & 92.8\%  & 91.4\%  & 99.4\%  & 98.7\%  \\
              & Wearing Necklace$^{0}$     & 71.8\%  & 71.4\%  & 86.9\%  & 84.2\%  \\
              & Wearing Necktie$^{+}$      & 83.7\%  & 82.1\%  & 98.5\%  & 98.0\%  \\ \rule{0pt}{3ex}\noindent
\parbox[t]{2mm}{\multirow{2}{*}{\rotatebox[origin=c]{90}{Other}}}         & Blurry$^{0}$                & 74.3\%  & 68.2\%  & 85.2\%  & 78.4\%  \\
              & Attractive$^{+}$            & 79.6\%  & 77.9\%  & 92.4\%  & 89.6\%  \\
\Xhline{2\arrayrulewidth}
\end{tabular}
\end{table}

\begin{table*}[]
\setlength{\tabcolsep}{3pt}
\renewcommand{\arraystretch}{1.0}
\centering
\caption{Prediction performance on LFW: the performance is based on FaceNet (FN) and ArcFace (AF) embeddings and is reported in terms of balanced accuracies at two difficulty scenarios: 100\% RCP (hard) and 50\% RCP (easy). $^{++}$,$^{+}$, and $^{0}$ state the assigned predictability class. Grey highlighting refers to reduced expressiveness due to limited data after the label-cleaning process.}
\label{tab:PerformanceLFW}
\begin{tabular}{ll}
\begin{tabu}[t]{llrrrr}
\Xhline{2\arrayrulewidth}
 & & \multicolumn{2}{c}{100\% RCP} & \multicolumn{2}{c}{50\% RCP} \\
 \cmidrule(rl){3-4} \cmidrule(rl){5-6}
 & Attribute & \multicolumn{1}{c}{FN}       & \multicolumn{1}{c}{AF}               & \multicolumn{1}{c}{FN} & \multicolumn{1}{c}{AF}         \\
\hline
\rule{0pt}{3ex}\noindent
\parbox[t]{2mm}{\multirow{8}{*}{\rotatebox[origin=c]{90}{Demographics}}}  & Male$^{++}$                         & 98.3\%       & 83.9\%       & 99.5\%        & 94.2\%       \\
         \rowfont{\color{gray}}     & Baby$^{0}$                         & 55.1\%       & 49.9\%       & 50.0\%        & 50.0\%       \\
              & Child$^{0}$                        & 68.8\%       & 57.5\%       & 75.8\%        & 52.4\%       \\
              & Youth$^{+}$                        & 79.9\%       & 70.5\%       & 93.1\%        & 79.8\%       \\
              & Middle Aged$^{+}$                 & 88.4\%       & 74.0\%       & 95.2\%        & 82.9\%       \\
              & Senior$^{++}$                       & 99.6\%       & 83.9\%       & 100.0\%       & 88.4\%       \\
              & Asian$^{++}$                        & 95.5\%       & 66.2\%       & 100.0\%       & 69.6\%       \\
              & White$^{++}$                        & 97.4\%       & 73.6\%       & 99.4\%        & 81.4\%       \\
              & Black$^{++}$                        & 95.3\%       & 63.2\%       & 98.3\%        & 53.6\%       \\
              & Indian$^{+}$                       & 85.2\%       & 50.2\%       & 92.5\%        & 54.7\%       \\
\rule{0pt}{3ex}\noindent
\parbox[t]{2mm}{\multirow{4}{*}{\rotatebox[origin=c]{90}{Skin}}}          & Rosy Cheeks$^{0}$                 & 67.2\%       & 58.8\%       & 73.0\%        & 64.3\%       \\
              & Shiny Skin$^{0}$                  & 82.1\%       & 67.9\%       & 89.7\%        & 75.6\%       \\
              & Pale Skin$^{0}$                   & 68.0\%       & 62.9\%       & 79.9\%        & 67.2\%       \\
              & Flushed Face$^{0}$                & 66.5\%       & 55.5\%       & 77.5\%        & 52.3\%       \\
\rule{0pt}{3ex}\noindent
\parbox[t]{2mm}{\multirow{6}{*}{\rotatebox[origin=c]{90}{Hairstyle}}}    
              & Curly Hair$^{0}$                  & 69.0\%       & 61.7\%       & 77.8\%        & 68.7\%       \\
              & Wavy Hair$^{++}$                   & 95.0\%       & 80.5\%       & 99.7\%        & 83.3\%       \\
              & Straight Hair$^{0}$               & 67.5\%       & 59.8\%       & 76.8\%        & 65.5\%       \\
              & Receding Hairline$^{+}$           & 83.3\%       & 73.0\%       & 93.5\%        & 84.9\%       \\
        \rowfont{\color{gray}}      & Bald$^{++}$                         & 93.6\%       & 75.8\%       & 97.9\%        & 75.0\%       \\
       \rowfont{\color{gray}}       & Bangs$^{++}$                        & 97.0\%       & 64.1\%       & 100.0\%       & 50.0\%       \\
              & Sideburns$^{++}$                    & 98.9\%       & 84.1\%       & 99.7\%        & 89.2\%       \\
\rule{0pt}{3ex}\noindent
\parbox[t]{2mm}{\multirow{3}{*}{\rotatebox[origin=c]{90}{Haircolor}}}      & Black Hair$^{++}$                  & 90.4\%       & 65.6\%       & 96.5\%        & 61.5\%       \\
              & Blond Hair$^{++}$                  & 95.2\%       & 71.7\%       & 98.8\%        & 55.6\%       \\
              & Brown Hair$^{+}$                  & 81.5\%       & 71.9\%       & 91.9\%        & 82.7\%       \\
              & Gray Hair$^{++}$                   & 98.8\%       & 88.4\%       & 100.0\%       & 93.9\%       \\
\rule{0pt}{3ex}\noindent
\parbox[t]{2mm}{\multirow{4}{*}{\rotatebox[origin=c]{90}{Beard}}}         & No Beard$^{++}$                    & 98.1\%       & 83.9\%       & 100.0\%       & 92.1\%       \\
              & Mustache$^{++}$                     & 98.5\%       & 79.7\%       & 99.3\%        & 78.1\%       \\
              & 5 o Clock Shadow$^{++}$          & 96.5\%       & 83.8\%       & 99.6\%        & 92.4\%       \\
         \rowfont{\color{gray}}     & Goatee$^{++}$                       & 94.5\%       & 70.0\%       & 100.0\%       & 100.0\%      \\
\rule{0pt}{3ex}\noindent
\parbox[t]{2mm}{\multirow{8}{*}{\rotatebox[origin=c]{90}{Face Geometry}}} & Oval Face$^{+}$                   & 82.7\%       & 71.6\%       & 95.4\%        & 75.8\%       \\
              & Square Face$^{++}$                 & 99.1\%       & 89.1\%       & 100.0\%       & 96.3\%       \\
           \rowfont{\color{gray}}   & Round Face$^{+}$                  & 84.2\%       & 49.6\%       & 100.0\%       & 50.0\%       \\
              & Round Jaw$^{0}$                  & 70.6\%       & 60.8\%       & 81.1\%        & 58.4\%       \\
        \rowfont{\color{gray}}       & Double Chin$^{++}$                 & 91.5\%       & 81.1\%       & 100.0\%       & 88.7\%       \\
              & High Cheekbones$^{+}$             & 79.9\%       & 73.3\%       & 90.4\%        & 81.8\%       \\
              & Chubby$^{+}$                       & 85.5\%       & 74.3\%       & 98.0\%        & 79.4\%       \\
              & Obstructed Forehead$^{+}$         & 85.9\%       & 65.0\%       & 99.9\%        & 61.3\%       \\
        \rowfont{\color{gray}}       & Partially Visible F.$^{+}$ & 85.2\%       & 65.9\%       & 94.0\%        & 50.0\%       \\
              & Fully Visible F.$^{+}$     & 85.9\%       & 71.8\%       & 95.4\%        & 82.2\%      \\
\Xhline{2\arrayrulewidth}
\end{tabu} 
&
\begin{tabu}[t]{llrrrr}
\Xhline{2\arrayrulewidth}
 & & \multicolumn{2}{c}{100\% RCP} & \multicolumn{2}{c}{50\% RCP} \\
 \cmidrule(rl){3-4} \cmidrule(rl){5-6}
 & Attribute & \multicolumn{1}{c}{FN}       & \multicolumn{1}{c}{AF}               & \multicolumn{1}{c}{FN} & \multicolumn{1}{c}{AF}         \\
\hline
\rule{0pt}{3ex}\noindent
\parbox[t]{2mm}{\multirow{5}{*}{\rotatebox[origin=c]{90}{Periocular}}}   & Eyes Open$^{0}$                & 60.4\% & 54.4\% & 63.6\%  & 54.8\% \\
            & Brown Eyes$^{+}$               & 82.1\% & 64.0\% & 92.8\%  & 66.8\% \\
            & Bags Under Eyes$^{+}$         & 87.2\% & 73.7\% & 95.4\%  & 83.5\% \\
            & Narrow Eyes$^{0}$              & 77.1\% & 66.2\% & 86.3\%  & 74.1\% \\
            & Bushy Eyebrows$^{++}$           & 96.3\% & 83.8\% & 99.1\%  & 91.7\% \\
            & Arched Eyebrows$^{+}$          & 85.3\% & 71.6\% & 94.5\%  & 76.8\% \\
\rule{0pt}{3ex}\noindent
\parbox[t]{2mm}{\multirow{5}{*}{\rotatebox[origin=c]{90}{Mouth}}}       & Mouth Closed$^{0}$             & 73.2\% & 64.0\% & 83.9\%  & 72.4\% \\
           \rowfont{\color{gray}} & Mouth Slightly Open$^{0}$     & 73.8\% & 61.8\% & 83.0\%  & 65.1\% \\
       \rowfont{\color{gray}}      & Mouth Wide Open$^{0}$         & 66.6\% & 50.8\% & 59.9\%  & 50.0\% \\
     \rowfont{\color{gray}}  & Teeth Not Visible$^{0}$       & 70.0\% & 65.2\% & 75.3\%  & 58.3\% \\
            & Smiling$^{0}$                   & 72.0\% & 67.9\% & 81.3\%  & 75.9\% \\
          \rowfont{\color{gray}}   & Big Lips$^{+}$                 & 87.6\% & 57.3\% & 98.0\%  & 57.8\% \\
\rule{0pt}{3ex}\noindent
\parbox[t]{2mm}{\multirow{3}{*}{\rotatebox[origin=c]{90}{Nose}}}         & Big Nose$^{+}$                 & 84.5\% & 71.6\% & 93.6\%  & 81.5\% \\
            & Pointy Nose$^{++}$              & 96.5\% & 71.5\% & 100.0\% & 71.3\% \\
            & Nose-Mouth Lines$^{0}$ & 70.0\% & 61.7\% & 80.7\%  & 71.6\% \\
\rule{0pt}{3ex}\noindent
\parbox[t]{2mm}{\multirow{7}{*}{\rotatebox[origin=c]{90}{Accessories}}} & Heavy Makeup$^{++}$             & 96.7\% & 69.9\% & 99.0\%  & 57.1\% \\
            & Wearing Hat$^{+}$              & 87.2\% & 67.9\% & 96.9\%  & 53.8\% \\
            & Wearing Earrings$^{++}$         & 91.7\% & 73.3\% & 97.9\%  & 72.9\% \\
            & Wearing Necktie$^{+}$          & 84.6\% & 72.8\% & 93.5\%  & 75.2\% \\
            & Wearing Necklace$^{+}$         & 83.7\% & 74.1\% & 92.1\%  & 82.5\% \\
            & Wearing Lipstick$^{++}$         & 98.5\% & 75.9\% & 99.5\%  & 74.0\% \\
            & No Eyewear$^{++}$               & 95.5\% & 86.1\% & 98.2\%  & 90.3\% \\
            & Eyeglasses$^{++}$                & 96.1\% & 90.0\% & 98.4\%  & 95.6\% \\
      \rowfont{\color{gray}}       & Sunglasses$^{0}$                & 71.6\% & 50.8\% & 62.4\%  & 50.0\% \\
\rule{0pt}{3ex}\noindent
\parbox[t]{2mm}{\multirow{4}{*}{\rotatebox[origin=c]{90}{Environment}}} & Blurry$^{0}$                    & 61.4\% & 57.2\% & 66.3\%  & 58.6\% \\
        \rowfont{\color{gray}}     & Harsh Lighting$^{0}$           & 76.0\% & 61.3\% & 89.1\%  & 57.9\% \\
            & Flash$^{0}$                     & 78.3\% & 58.3\% & 88.3\%  & 51.5\% \\
            & Soft Lighting$^{0}$            & 65.7\% & 60.2\% & 72.3\%  & 66.1\% \\
    \rowfont{\color{gray}}         & Outdoor$^{0}$                   & 77.2\% & 60.8\% & 81.9\%  & 65.9\% \\
\rule{0pt}{3ex}\noindent
\parbox[t]{2mm}{\multirow{4}{*}{\rotatebox[origin=c]{90}{Other}}}       & Frowning$^{0}$                  & 78.3\% & 72.4\% & 88.8\%  & 79.5\% \\
          \rowfont{\color{gray}}  & Color Photo$^{0}$              & 72.8\% & 54.0\% & 75.0\%  & 60.0\% \\
            & Posed Photo$^{0}$              & 76.0\% & 60.7\% & 80.9\%  & 63.0\% \\
            & Attractive Man$^{0}$           & 74.4\% & 65.0\% & 85.1\%  & 74.2\% \\
            & Attractive Woman$^{++}$         & 95.3\% & 75.1\% & 100.0\% & 71.4\% \\
\Xhline{2\arrayrulewidth}
\end{tabu}
\end{tabular}
\end{table*}

To derive statements of which attributes are encoded in face templates, the prediction performance of the attributes is determined at two difficulty-levels.
100\% RCP (hard) refers to the use of all samples under the given circumstances.
50\% RCP (easy) refers to the 50\% the predictions of which the classifier is most sure about its correctness.
In Table \ref{tab:PerformanceCelebA} the prediction performance is shown for CelebA including the assigned predictability classes.
Tow general observations are made. 
First, the performance at the 50\% RCP-level is always higher than for 100\% RCP showing that MAC learned reliable predictions on the dataset.
Second, even if the prediction performance on FaceNet (FN) and ArcFace (AF) is very similar, the performance on FN is always slightly higher.
This can be explained by ArcFace's margin-principle during training that distorts the feature space more incoherently and thus, makes it harder for pattern learning.
In total, many of CelebA attributes can be predicted with high accuracy from face templates.
This includes demographic characteristics such as gender, characteristics of the person's hairstyle, haircolor, and about the beard.
Moreover, the deeply encoded features also contain highly-detailed information about the person's accessories.

\vspace{-1mm}
Table \ref{tab:PerformanceLFW} shows the same evaluation setting on the LFW database.
The grey highlights refer to results with limited significance since the label-cleaning process eliminated many samples with low-quality labels.
The low number of train- and testing-samples explains some of the weak performance such as for \textit{Baby}, \textit{Sunglasses}, and the \textit{Mouth} category.
However, comparing the results of LFW with the results of CelebA (Table \ref{tab:PerformanceCelebA}) shows similar performances on attributes which occur in both datasets, such as demographic attributes, haircolors, face geometry etc. 
Consequently, our label-cleaning process removed low-quality attribute-labels but did not result in a large bias of the data.
Due to the entangled patterns encoded in the templates some attributes, such as \textit{Bold}, \textit{Bangs}, and \textit{Goatee}, are easy to learn and thus, achieve high performances.
Generally, the prediction performance using ArcFace embeddings is significantly weaker than using FaceNet.
ArcFace embeddings contain more complex attribute patterns and for the experiments on LFW less data was available for training, since we manually filtered low-quality labels.
Consequently, it can be expected that with more training data the performance on ArcFace is higher.
Nevertheless, similar to CelebA, many attributes can be predicted with high accuracies from the templates only.
This goes for demographic attributes such as gender, age, and race, as well as for hairstyle, haircolor, beard, and accessories.
Moreover, characteristics about the face geometry such as face shape, double chin, and forehead visibility can be determined.
Factors that do not belong to the person, such as lighting conditions and blurriness, can not be predicted reliably with the MAC.
It is interesting to note that the high predictability of \textit{Attractive woman} can be explained by the high correlation to accessories.

\vspace{-1mm}
\subsection{Summary}
\label{sec:Summary}
\vspace{-2mm}

From 113 investigated attributes, we found that 39 attributes belong to easily-predictable, 35 belong to predictable and 39 to hardly-predictable.
To obtain a more general overview of the encoded information in face templates, Table \ref{tab:Summary} summarizes the categories of the attributes in the three predictability classes.
The assignment of the categories to the individual attributes is shown in Table \ref{tab:PerformanceLFW}.
Providing a more complete view of the problem, this table also includes findings from related works.
Since the face templates are trained with the purpose of recognition, it seems logical that categories such as \textit{Face Geometry}, \textit{Periocular Area}, \textit{Nose}, and \textit{Mouth} are easily-predictable.
Surprisingly, this is not the case.
Instead, non-permanent factors such as \textit{Hairstyle}, \textit{Haircolor}, \textit{Beard}, \textit{Accessories}, \textit{Head Pose}, and \textit{Social Traits} are easily-predictable.
Modern face recognition systems aim to be robust against these factors and still these factors are strongly present in face templates.

For many applications, the user of a face recognition system solely provides his biometric data for recognition. 
To prevent a function creep of his data, face templates should contain only identity-related information.
However, the experiment showed that many privacy-sensitive attributes are encoded in face templates.
This raises a major privacy risk.
Consequently, future works might analyse the reason for this rich encodings and find solutions to preserve privacy in face recognition systems.

\begin{table}[H]
\setlength{\tabcolsep}{3pt}
\renewcommand{\arraystretch}{1.0}
\centering
\caption{Categorized summary of the predictability classes including findings of related works.}
\label{tab:Summary}
\begin{tabular}{lll}
\Xhline{2\arrayrulewidth}
Easily-predictable  & Predictable         & Hardly-predictable       \\
\hline
Demographics     & Face Geometry & Skin            \\
Hairstyle        & Periocular     & Mouth           \\
Haircolor        & Nose          & Environment     \\
Beard            & Image Quality \cite{DBLP:journals/corr/Best-RowdenJ17}              &            \\
Accessories      &               &              \\   
Head Pose \cite{DBLP:conf/fgr/PardeCHCSCO17} & & \\
Social Traits \cite{DBLP:journals/cogsci/PardeHCSO19} & & \\
\Xhline{2\arrayrulewidth}              
\end{tabular}
\end{table}
% put ser fiq here

%%%%%%%%% Conclusion
\vspace{-5mm}
\section{Conclusion}
\label{sec:Conclusion}
% written for the read who now knows about the paper
% what can the reader do with the newly acquired knowledge?

% 1 - underline the importance of the topic
The success of current face recognition systems is based on the advances of deeply-learned templates.
%These templates are learned to recognize individuals despite daily variations.
% 2 - (revisit previous research and) recall specific weakness in the methodology used in previous studies
Recent works have shown that demographics, image characteristics, and social traits are encoded in these templates.
This can lead to biased decisions in face recognition systems and raises major privacy issues.
In many applications these templates are expected to be used for recognition purposes only and deducing information that is not required for recognition is considered as a violation of their privacy.
% 3 - revisit the methodology used in this study
The knowledge of the encoded information in face templates is necessary to develop effective bias-mitigating and privacy-preserving technologies.
The main contribution of this work is an analysis of what information is stored in face templates.
% 4 - revisit and summarise the results
More precisely, 113 attributes are analyses towards their predictability from face templates.
The experiments were conducted on two popular face templates under two difficulty-levels.
% to make more sophisticated statements about the attribute predictability.
To facilitate the understandability of the results, each attribute was further categorized into one of three predictability classes.
% 5 - show where and how the present work fits into the research "map" in this field
% 6 - recalls an positive achievement/contribution of this work
Results reveal that about one third of the analysed attributes are easily-predictable, another third is predictable, and one third is hardly-predictable.
%Results reveal that 39 and 74 attributes belong to the easily-predictable and predictable classes showing that these characteristics are accurately predictable under challenging and close-to-optimal difficulty settings.
% 7 - focus on the meaning and implications of the achievements in this work
Despite that face recognition templates are trained to be robust against non-permanent factors, the results demonstrate that especially these attributes are accurately predictable from face templates.
%This includes information about age, hairstyles, haircolors, beards, and various accessories.
% 8 - note the one of the achievemens /contributions of this work is novel
% 9 - describe future research directions
Future works might build on the knowledge of this work to develop comprehensive bias-mitigating and privacy-preserving solutions for face recognition.

%\section*{Acknowledgments}
%This research work has been funded by the German Federal Ministery of Education and Research and the Hessen State Ministry for Higher Education, Research and the Arts within their joint support of the National Research Center for Applied Cybersecurity. Portions of the research in this paper use the FERET database of facial images collected under the FERET program, sponsored by the Counterdrug Technology Development Program Office.

\clearpage
\newpage

\section*{Acknowledgement}
This work was supported by the German Federal Ministry of Education and Research (BMBF) as well as by the Hessen State Ministry for Higher Education, Research and the Arts (HMWK) within the National Research Center for Applied Cybersecurity (ATHENE), and in part by the German Federal Ministry of Education and Research (BMBF) through the Software Campus project. 

{\small
\bibliographystyle{ieee}
\bibliography{egbib}
}

\end{document}